\documentclass[nohyperref]{article}

\usepackage{amsmath,amsfonts,bm, amsthm, mathrsfs, mathtools}

\newtheorem{theorem}{Theorem}

\newtheorem{proposition}{Proposition}
\newtheorem{remark}{Remark}

\newtheorem{definition}{Definition}

\numberwithin{theorem}{section}
\numberwithin{lemma}{section}
\numberwithin{proposition}{section}
\numberwithin{corollary}{section}

\def\1{\bm{1}}

\def\vd{{\bm{d}}}

\def\vf{{\bm{f}}}
\def\vg{{\bm{g}}}
\def\vh{{\bm{h}}}

\def\vp{{\bm{p}}}

\def\vw{{\bm{w}}}
\def\vx{{\bm{x}}}

\DeclareMathAlphabet{\mathsfit}{\encodingdefault}{\sfdefault}{m}{sl}
\SetMathAlphabet{\mathsfit}{bold}{\encodingdefault}{\sfdefault}{bx}{n}

\def\gD{{\mathcal{D}}}

\def\gH{{\mathcal{H}}}

\def\gL{{\mathcal{L}}}

\def\gP{{\mathcal{P}}}

\def\gX{{\mathcal{X}}}
\def\gY{{\mathcal{Y}}}
\def\gZ{{\mathcal{Z}}}

\def\sR{{\mathbb{R}}}

\newcommand{\E}{\mathbb{E}}

\DeclareMathOperator*{\argmax}{arg\,max}
\DeclareMathOperator*{\argmin}{arg\,min}

\usepackage{microtype}
\usepackage{graphicx}
\usepackage{subfigure}
\usepackage{booktabs} 

\usepackage{hyperref}
\usepackage{url}

\newcommand{\nn}{\nonumber}

\usepackage{amsfonts}       
\usepackage{mathrsfs}
\usepackage{nicefrac}       
\usepackage{xcolor}         
\usepackage{enumitem}
\usepackage{quoting}
\usepackage{wrapfig}
\usepackage{makecell}
\usepackage{bbm}
\usepackage{dsfont}
\usepackage{caption} 
\usepackage{comment}
\usepackage[ruled,algo2e]{algorithm2e}
\usepackage{algorithm}

\usepackage[accepted]{icml2023}
\hypersetup{
    colorlinks=true,
    citecolor =cyan,
    linkcolor=magenta,
    urlcolor=blue,
}

\usepackage{amsmath}
\usepackage{amssymb}
\usepackage{mathtools}
\usepackage{amsthm}

\usepackage[textsize=tiny]{todonotes}

\icmltitlerunning{On Balancing Bias and Variance in MSFDA}

\begin{document}

\twocolumn[
\icmltitle{On Balancing Bias and Variance in\\ Unsupervised Multi-Source-Free Domain Adaptation}
\icmlsetsymbol{equal}{*}

\begin{icmlauthorlist}
\icmlauthor{Maohao Shen}{mit}
\icmlauthor{Yuheng Bu}{uf}
\icmlauthor{Gregory Wornell}{mit}
\end{icmlauthorlist}

\icmlaffiliation{mit}{Department of Electrical Engineering and Computer Science, Massachusetts Institute of Technology, Cambridge, USA}
\icmlaffiliation{uf}{Department of Electrical \& Computer Engineering, University of Florida, Gainesville, USA}
\icmlcorrespondingauthor{Maohao Shen}{maohao@mit.edu}

\icmlkeywords{Machine Learning, ICML}

\vskip 0.3in
]



\printAffiliationsAndNotice{}  

\begin{abstract}
Due to privacy, storage, and other constraints, there is a growing need for unsupervised domain adaptation techniques in machine learning that do not require access to the data used to train a collection of source models. Existing methods for multi-source-free domain adaptation (MSFDA) typically train a target model using pseudo-labeled data produced by the source models, which focus on improving the pseudo-labeling techniques or proposing new training objectives. Instead, we aim to analyze the fundamental limits of MSFDA. In particular, we develop an information-theoretic bound on the generalization error of the resulting target model, which illustrates an inherent bias-variance trade-off. We then provide insights on how to balance this trade-off from three perspectives, including domain aggregation, selective pseudo-labeling, and joint feature alignment, which leads to the design of novel algorithms. Experiments on multiple datasets validate our theoretical analysis and demonstrate the state-of-art performance of the proposed algorithm, especially on some of the most challenging datasets, including Office-Home and DomainNet.

\end{abstract}

\section{Introduction}  \label{sec:intro}

Machine learning models trained in a standard supervised manner suffer from the problem of domain shift~\citep{quinonero2008dataset}, i.e., directly applying the model trained on the source domain to a distinct target domain usually results in poor generalization performance. Unsupervised Domain Adaptation (UDA) techniques have been proposed to mitigate this issue by transferring the knowledge learned from a labeled source domain to an unlabeled target domain. One prevailing UDA strategy to resolve the domain shift issue is domain alignment, i.e., learning domain-invariant features either by minimizing the discrepancy between the source and target data ~\citep{long2015learning, long2018transferable, peng2019moment} or through adversarial training~\citep{ganin2015unsupervised, tzeng2017adversarial}. However, traditional UDA methods require access to labeled source data and only apply to the single source domain adaptation, which cannot fulfill the emerging challenges in real-world applications.

In practice, the source data might not be available due to various reasons: privacy preservation, i.e., the data that contains sensitive information, such as health and financial status, is unsuitable to be made public; and storage limitations, i.e., large-scale datasets, such as high-resolution videos, require substantial storage space. Due to these practical concerns, the source-free domain adaptation (SFDA) problem has attracted increasing attentions~\citep{yang2020unsupervised, kim2020domain, liang2020we, li2020model}, which aims to address the data-free challenge by adapting the pretrained source model to the unlabeled target domain. 

The other practical concern is that the source data is usually collected from multiple domains with different underlying distributions, such as the street scene from different cities~\citep{cordts2016cityscapes} and the biomedical images with different modalities~\citep{dou2018unsupervised}. Taking such practical consideration into account, multi-source domain adaptation (MSDA)~\citep{guo2018multi, peng2019moment} aims to adapt to the target domain by properly aggregating the knowledge from multiple source domains.

A more challenging scenario is to combine both the data-free and multi-source settings, i.e., the source data collected from multiple domains with distinct distributions is not accessible due to some practical constraints. For example, federated learning~\citep{truong2021privacy} aggregates the information learned from a group of heterogeneous users. To preserve user privacy, the data of each user is stored locally, and only the trained models are transmitted to the central server. 

\begin{figure*}[!t]
  \centering
  \captionsetup{font=small}
  \includegraphics[width=0.86\linewidth]{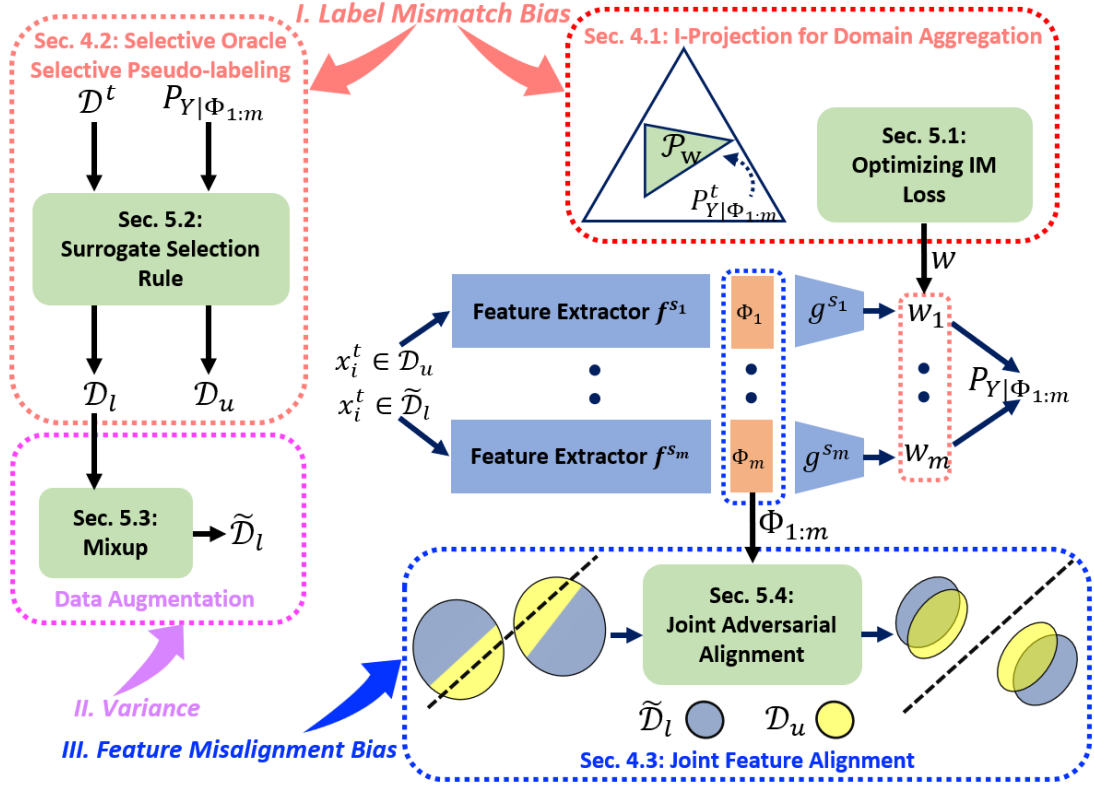}
  \caption{Schematic of our theoretical insights and algorithm design. More details on label mismatch bias, feature misalignment bias, and variance can be found in the corresponding sections, as shown in the figure.}
  \label{fig:schematic}
\end{figure*}

We consider the Multi-Source-Free Domain Adaptation (MSFDA) problem to overcome these two challenges. The MSFDA problem is less explored, and few methods have been proposed. Existing methods either propose new training objectives motivated by specific intuitions without a theoretical justification~\citep{yang2021exploiting, kundu2022balancing}, or leverage on generating pseudo-labels using source models but not fully address the bias induced by the noisy pseudo-labeling procedure~\citep{ahmed2021unsupervised, dong2021confident}. In addition, unlike traditional domain adaptation approaches, none of these methods can explicitly address the domain shift issue in source and target domains. To this end, we aim to understand the fundamental limit of the MSFDA problem through theoretical analysis and then draw some insights for new algorithm design. We show that three factors control the generalization error of the MSFDA problem: \textit{label mismatch bias}, \textit{feature misalignment bias}, and \textit{variance} depending on the number of training samples. As in Figure~\ref{fig:schematic}, we demonstrate how to balance the trade-off by providing three crucial insights: i.e., (1) Appropriate domain aggregation of multiple source domains can reduce the \textit{label mismatch bias}, (2) selective pseudo-labeling a subset of data for training can further balance the \textit{label mismatch bias} and the variance, and (3) Utilizing a joint feature alignment strategy to explicitly address the domain shift issue by reducing the \textit{feature misalignment bias}.

\textbf{Summary of contributions:} (1)We develop an information-theoretic generalization error bound for the MSFDA problem, demonstrating an inherent bias-variance trade-off. (2) We provide theoretical understandings and draw insights into how to balance the bias and variance trade-off from three perspectives. (3) Motivated by our theoretical analysis, we empirically study the performance limit of the MSFDA problem by providing a performance upper bound and propose a novel algorithm. (4) Experiments across multiple representative benchmark datasets validate our theoretical results and demonstrate the superior performance of the proposed algorithm over existing methods.

\section{Related Work}   \label{sec: related}



\textbf{Multi-source Domain Adaptation:}
Multi-source domain adaptation aims to transfer knowledge from multiple distinct source domains to a target domain. Early theoretical works provide theoretical guarantees for later empirical works by formally characterizing the connection between the source and target domains.  \citep{ben2010theory} introduces the $\gH \Delta \gH$ distance to measure the discrepancy between the target  and source domains, and \citep{mansour2009domain} assumes that the target distribution can be approximated by a linear combination of source distributions. 
Many existing algorithms aim to mitigate the distribution shift issue between source and target domains. Discrepancy-based methods try to align the domain distribution by minimizing discrepancy loss, such as maximum mean discrepancy (MMD)~\citep{guo2018multi}, Rényi-divergence~\citep{hoffman2018algorithms}, moment distance~\citep{peng2019moment}, and a combination of different discrepancy metrics~\citep{guo2020multi}. Adversarial methods align the features between source and target domains by training a feature extractor that fools the discriminator under different loss functions, including $\gH$-divergence~\citep{zhao2018adversarial}, traditional GAN loss~\citep{xu2018deep}, and Wasserstein distance~\citep{li2018extracting, wang2019tmda, zhao2020multi}. Moreover, the domain weighting strategy is also widely used to quantify the contribution of each source domain, including uniform weights~\citep{zhu2019aligning}, source model accuracy-based weights~\citep{peng2019moment}, Wasserstein distance-based weights~\citep{zhao2020multi}, and aggregating multiple domains using graph model~\citep{wang2020learning}

\textbf{Source-free Domain Adaptation:}
All the methods mentioned above require source data and cannot be directly applied to the data-free setting, and recent efforts have been made to tackle the source-free problem. \citep{xia2021adaptive} constructs a target-specific classifier to mitigate the domain discrepancy.  Adversarial learning-based methods utilize generative models by either generating new samples from the source domain~\citep{kurmi2021domain} 
or generating labeled samples from target distribution by conditional GAN~\citep{li2020model}. Another strategy is using the pseudo-labeling technique. \citep{liang2020we} uses self-supervised pseudo-labeling and maximizes the mutual information loss. Similarly, \citep{kim2020domain} proposes a confidence-based filtering method to further improve the quality of pseudo-labels. 

\textbf{Multi-source-free Domain Adaptation:}
To overcome both data-free and multi-source challenges, \citep{ahmed2021unsupervised} adapts the self-supervised pseudo-labeling method in \citep{liang2020we} to the multi-source setting. Similarly, \citep{dong2021confident} also focuses on improving the pseudo-labeling algorithm with a confident-anchor-induced pseudo-label generator. Other works focus on designing new training objectives, including~\citep{yang2021exploiting, yang2021generalized} that proposes a nearest-neighbor-based regularizer to encourage prediction consistency, and~\citep{kundu2022balancing} that designs new data augmentation techniques to balance the discriminability and transferability trade-off for source free domain adaptation. In contrast to many prior works that are often ad-hoc and lack theoretical justification, our work aims to design novel MSFDA algorithms by analyzing the MSFDA problem using information-theoretical tools. While  \citep{dong2021confident} also conducts theoretical analysis, their results require specific assumptions, such as the meta-assumption of the data distribution. Our theoretical result is more general and does not require additional assumptions regarding the data distribution or models.

\section{Problem Formulation}   \label{sec: notation}
Let $\gZ = \gX \times \gY$, where $\gX$ denotes the input space and $\gY=\{1,\cdots,K\}$ denotes the label space. A domain is defined by a joint distribution $P_{XY}$ on the instance space $\gZ$.   In this work, we aim to jointly adapt multiple pretrained models corresponding to $m$ different source domains $\{P^{s_j}_{XY}\}_{j=1}^m$ to a new target domain $P^{t}_{XY}$. 
Let $\vh^{s_j}:\gX \rightarrow \Delta^{K-1}$ denote the pretrained model for source domain $j$, which is a function predicting the conditional distribution $P^{s_j}_{Y|X}$ in probability simple $\Delta^{K-1}$, i.e., $\vh^{s_j} = [h^{s_j}_1,\cdots,h^{s_j}_K]^\top$, $\sum_{k=1}^K h^{s_j}_k =1$ and $h^{s_j}_k\ge 0$. Each pretrained model can be decomposed into a feature extractor $\vf^{s_j}: \gX \rightarrow \mathcal{F}_j$, followed by a classifier $\vg^{s_j}: \mathcal{F}_j \rightarrow \Delta^{K-1}$, where $\mathcal{F}_j$ denotes the representation space of each source model. Thus, we can denote the prediction of input $\vx$ using model $\vh^{s_j}$ as $\vh^{s_j}(\vx) =\left(\vg^{s_j} \circ \vf^{s_j}\right)(\vx)$. For any  feature representations $\phi_j \in \mathcal{F}_j$, each classifier $\vg^{s_j}(\bm{\phi}_j)$ induces a conditional distribution $P^{s_j}_{Y|\Phi_j=\phi_j}$.

Denote $\gD^{t}\triangleq\{\vx^t_i\}_{i=1}^{n}$ as the unlabeled target domain dataset, where $\vx^t_i$ are i.i.d. generated from the target marginal distribution $P^t_X$. For any $\vx_i^t\in \gD^{t}$, we denote $\bm{\phi}^{s_j}_i = \vf^{s_j}(\vx_i^t)$ as the feature representation of sample $i$ based on source model $s_j$. We aim to construct a target model $\vh$ that aggregates information from multiple source models by modifying the feature mappings $\vf^{s_j}$. Denote the final prediction as $\hat{y} =\vh\left(\{\vf^{s_j}(\vx) \}_{j=1}^m  \right) = \vh\left(\{\bm{\phi}_j \}_{j=1}^m\right)$. For the loss function $\ell: \gY \times \gY \to \sR_+$, our ultimate goal is to obtain a target model that minimizes the population risk of the target domain, i.e.,  $\gL_P\left(\vh,P^t_{XY}\right)\triangleq\E_{P^t_{XY}} [\ell\left(\vh\left(\{\vf^{s_j}\left(X\right)\}_{j=1}^m  \right), Y\right)]$.



\section{Theoretical Analysis and Insights}   \label{sec: theorem}

In this section, we provide a theoretical analysis of the multi-source-free domain adaptation problem. We show that there exists an inherent bias and variance trade-off that needs to be balanced in our algorithm design. Suppose that for some unlabeled target samples $\vx^t_i \in \gD^t$, we can obtain its pseudo-label $\tilde{y}_i^t$ by leveraging the pretrained models, and denote the subset of pseudo-labeled data as 
 $\gD_{l}\triangleq\{(\vx_i^t,\tilde{y}_i^t )\}_{i=1}^{n_l}$. To ensure that model learned by minimizing the empirical risk over  $\gD_{l}$, i.e.,
\begin{equation}
    \gL_E(\vh,\gD_l) \triangleq \frac{1}{n_l}\sum_{i=1}^{n_l} \ell \left(\vh\left(\{\bm{\phi}^{s_j}_i\}_{j=1}^m\right), \tilde{y}^t_i\right)
\end{equation} 
generalizes well to the target domain, we have the following upper bound on generalization error, i.e., the difference between the  population risk and the empirical risk. 

\begin{theorem}\label{thm:gen_bound}(proved in Appendix \ref{appendix: proof})
Suppose that the samples of $\gD_l$ are i.i.d. generated from the distribution $P^{\gD_l}_{XY}$, the function space $\gH$ has finite Natarajan dimension $d_N(\gH)$, then for any loss function bounded in $[0, M]$ and  any $\vh\left(\cdot\right) \in \gH $, there exists a constant $C$ such that with probability $1-\delta$,
\begin{align}\label{equ: gen_bound}
     &\gL_P(\vh,P^t_{XY}) - \gL_E(\vh,\gD_l) \nonumber \\ 
     \le &  \frac{M}{\sqrt{2}}\sqrt{ \underbrace{D(P^{\gD_l}_{Y|\Phi_{1:m}}\|P^t_{Y|\Phi_{1:m}} |P^{\gD_l}_{\Phi_{1:m}} )}_{\text{Bias: label distribution mismatch}} +
     \underbrace{D(P^{\gD_l}_{\Phi_{1:m}}\|P^t_{\Phi_{1:m}})}_{\text{Bias: feature misalignment}}} \nonumber\\
     &+C\underbrace{\sqrt{\frac{d_N(\gH)\log K + \log\frac{1}{\delta} }{n_l}}}_{\text{Variance: number of labeled sample}},
\end{align}
where $\Phi_{1:m} \triangleq \{\vf^{s_j}(X) \}_{j=1}^m$ are random vectors induced by different features mappings $\vf^{s_j}: \gX \rightarrow \mathcal{F}_j$.
\end{theorem}
Theorem~\ref{thm:gen_bound} states that the generalization error can be controlled by a bias and variance trade-off. The first term can be viewed as bias, where KL divergence measures the discrepancy between two distributions. The first term can be further decomposed into two sub-terms: the bias due to the mismatch between the pseudo label distribution and the target label distribution; and the bias due to the misalignment of marginal feature distribution in the representation space. The second term in \eqref{equ: gen_bound} can be interpreted as the variance since it only depends on the hypothesis space $\gH$ and the number of the pseudo-labeled samples $n_l$.

In the following, we discuss how to balance the bias-variance trade-off by improving each term in the generalization error bound provided in Theorem~\ref{thm:gen_bound}. First, we make the connection between minimizing the \textit{label mismatch bias} and domain aggregation of multiple source domains. Second, we reveal that selective pseudo-labeling is essential to prevent the 
\textit{label mismatch bias} from being unbounded while also balancing the variance term. Finally, we show how a joint feature alignment approach is crucial for the MSFDA problem to reduce the \textit{feature misalignment bias}. 

\subsection{Domain Aggregation} \label{sec:thm-aggregation}
To leverage the pretrained source models and construct a labeling distribution $P^{\gD_l}_{Y|\Phi_{1:m}}$ that reduces the \textit{label mismatch bias}  $D(P^{\gD_l}_{Y|\Phi_{1:m}}\|P^t_{Y|\Phi_{1:m}} |P^{\gD_l}_{\Phi_{1:m}} )$, we need to aggregate the information from multiple source models. One naive approach is to weight the predictions from multiple models $P^{s_j}_{Y|\Phi_j}$ uniformly~\cite {zhu2019aligning}. However, each source domain may have different transferability, and treating them equally is sub-optimal. Here, we consider a mixture distribution with non-negative domain weights $\vw$, i.e.,  $P^{w}_{Y|\Phi_{1:m}} = \sum_{j=1}^m \vw_j P^{s_j}_{Y|\Phi_j}$ to aggregate multiple source models. We denote the convex set of such mixture distribution as $\gP_{w} = \{q~|~ q=\sum_{j=1}^m \vw_j P^{s_j}_{Y|\Phi_j}, \ \sum_{j=1}^m \vw_j=1\}$. In order to minimize the \textit{label mismatch bias}, the pseudo label distribution $P^{\gD_l}_{Y|\Phi_{1:m}}$ should minimize its KL divergence to the target label distribution $P^{t}_{Y|\Phi_{1:m}}$, i.e., 
\begin{equation} \label{eq:i-projection}
    P^{\gD_l}_{Y|\Phi_{1:m}} = \argmin_{P\in \gP_w}D\left( P \| P^{t}_{Y|\Phi_{1:m}}\right).
\end{equation}
Notice that this is equivalent to let $P^{\gD_l}_{Y|\Phi_{1:m}}$ to be the \textit{I-projection}~\citep{csiszar1984sanov} of target distribution onto $\gP_w$, where we have the following proposition,
\begin{proposition}\label{proposition:mixture}(proved in Appendix \ref{appendix:proposition})
Let $\{P^{s_j}_{Y|\Phi_j}\}_{j=1}^m$ be a collection of source models, and let $P^{t}_{Y|\Phi_{1:m}}$ be any target label distribution, then there exists a mixture model with weights $\vw^*$ such that 
\begin{equation*}
D\Big( \sum_{j=1}^m \vw^*_j P^{s_j}_{Y|\Phi_j} \| P^{t}_{Y|\Phi_{1:m}}\Big) \leq D\big(P^{s_j}_{Y|\Phi_j}  \| P^{t}_{Y|\Phi_{1:m}}\big)
\end{equation*}
holds for any $j$.
\end{proposition}
Proposition~\ref{proposition:mixture} states that the mixture of source models can induce lower bias compared to using any single model, which justifies the benefits of aggregating the multiple source models to reduce the \textit{label mismatch bias}. Besides, as the mixture distribution defines a convex hull over the probability simplex, increasing the number of source models will neither shrink the convex hull nor raise the approximation error. Therefore, utilizing more source domains with the optimal mixture weights $\vw^*$ will benefit the domain adaptation problem. In practice, the target label distribution $P^{t}_{Y|\Phi_{1:m}}$ in~\eqref{eq:i-projection} is not available, and we discuss how to approximate the mixture distribution in Section~\ref{sec:aggregation}.

\subsection{Selective Pseudo-labeling}  \label{sec:pseudo-labeling}
Once we obtain the pseudo labeling distribution $P^{\gD_l}_{Y|\Phi_{1:m}}$, the next step is to generate pseudo-labels using this distribution. Most existing work generate pseudo-labels for all samples in $\gD^t$, and the joint distribution of $\gD_l$ is given by $P^{\gD_l}_{\Phi_{1:m},Y} = P_{\Phi_{1:m}}^t \otimes P^{\gD_l}_{Y|\Phi_{1:m}}$. Although generating pseudo-labels for the entire $\gD^t$ implies $n_l=n$, which reduces the variance term in Theorem~\ref{thm:gen_bound},  the \textit{label mismatch bias} might be unbounded due to the approximation error in equation~\ref{eq:i-projection}. To see this, if $P^t_{Y|\Phi_{1:m}}(Y=y|\{\Phi_j=\bm{\phi}^{s_j}_i\}_{j=1}^m)=0$ and  $P^{\gD_l}_{Y|\Phi_{1:m}}(Y=y|\{\Phi_j=\bm{\phi}^{s_j}_i\}_{j=1}^m)\ne 0$ for some $\vx_i \in \gD^t$ due to model mismatch, it leads to a large \textit{label mismatch bias}, i.e., $D(P^{\gD_l}_{Y|\Phi_{1:m}}\|P^t_{Y|\Phi_{1:m}}|P^{\gD_l}_{\Phi_{1:m}})=\infty$. 

The aforementioned issue can be mitigated by only generating pseudo-labels for a subset of $\gD^t$. We start our discussion by assuming the following selective oracle, which gives a perfect subset selection. Such an assumption helps us to understand the performance limits of the MSFDA problem revealed in Theorem~\ref{thm:gen_bound}, which is further validated by the empirical results provided in~\ref{sec:exp-results-oracle}.
\begin{definition}\label{def:oracle}(Selective Oracle)
Given the labeling distribution for target domain data $P^{\gD_l}_{Y|\Phi_{1:m}}(\cdot|\cdot)$, the selective oracle identifies a subset of data $\gX_{\gD_l}$ for pseudo labeling, which satisfies the following criterion:
\begin{align} \label{equ:definition}
    &P^{\gD_l}_{Y|\Phi_{1:m}}(\cdot|\{\Phi_j=\vf^{s_j}(\vx)\}_{j=1}^m) \nonumber \\
    &= P^t_{Y|\Phi_{1:m}}(\cdot|\{\Phi_j=\vf^{s_j}(\vx)\}_{j=1}^m),
    \forall\vx \in\gX_{\gD_l}.
\end{align}
\end{definition}
With the selective oracle, the \textit{label mismatch bias} can be bounded and reduced to a negligible value. Note that the variance term is determined by the cardinality of subset $\gX_{\gD_l}$, i.e., $n_l = |\gX_{\gD_l}|$. The oracle induces a small variance term by selecting a sufficient number of data in $\gX_{\gD_l}$ while also ensuring a bounded \textit{label mismatch bias}. In practice, when such an oracle is not available, we propose a surrogate selection technique in~\ref{sec:surrogate} to balance such a trade-off. 

\subsection{Joint Feature Alignment} \label{sec:thm-alignment}
Notice that there exists the distribution shift between data marginal distributions in the joint representation space for $\gD_l$ and $\gD_t$, i.e., $P^{\gD_l}_{\Phi_{1:m}} \neq P^{\gD_t}_{\Phi_{1:m}}$, where we have the following remark about the \textit{feature misalignment bias}.
\begin{remark} (proved in Appendix \ref{appendix:remark}) \label{remark:joint-align}
The \textit{feature misalignment bias} in the joint representation space induced by multiple source models is the upper bound of the average \textit{feature misalignment bias} across each representation space, i.e.,
\begin{equation}
D(P^{\gD_l}_{\Phi_{1:m}}\|P^t_{\Phi_{1:m}}) \geq \frac 1 m \sum_{j=1}^{m} D(P^{\gD_l}_{\Phi_j}\|P^t_{\Phi_{j}}).
\end{equation}
\end{remark}
\vspace{-1em}
Remark \ref{remark:joint-align} states that $D(P^{\gD_l}_{\Phi_{1:m}}\|P^t_{\Phi_{1:m}})=0$ ensures $\sum_{j=1}^{m} D(P^{\gD_l}_{\Phi_j}\|P^t_{\Phi_{j}}) =0$, but not vice versa. This highlights that the feature alignment needs to be conducted in the joint representation space instead of in each source domain separately to further reduce the \textit{feature misalignment bias}, which motivates us to propose a joint adversarial feature alignment loss in~\ref{sec:alignment}.

\section{Practical Algorithm}    \label{sec:method}
The theoretical analysis presented above sheds light on how to balance the bias and variance terms as revealed by the generalization error bound in Theorem~\ref{thm:gen_bound}. However, in practice, the target label information used in the previous analysis is not available. To this end, we introduce practical algorithms that incorporate these insights. First, we discuss methods for learning domain aggregation weights without target domain labels. Then, we propose a surrogate selection rule for the selective Oracle and introduce a joint adversarial feature alignment approach. Finally, we present an overall algorithm for solving the MSFDA problem.

\subsection{Learning Domain Aggregation Weights} \label{sec:aggregation}
Obtaining the optimal labeling distribution $P^{\gD_l}_{Y|\Phi_{1:m}}$ through \textit{I-projection} (equation~\ref{eq:i-projection}) is not possible as the ground-truth target distribution is not accessible. However, we can make the following assumptions about the optimal mixture distribution. In general, the predictions of target data should be individually confident and globally diverse. First, for any input data $\vx_i^t \in \gD^t$, the mixture labeling distribution $\sum_{j=1}^m \vw_j P^{s_j}_{Y|\Phi_j}(\cdot|\Phi_j=\bm{\phi}^{s_j}_i)$ should be confident and lie in one of the corners of the probability simplex. Thus, it is appropriate to minimize the entropy of the mixture labeling distribution of individual input data. Second, the marginal label distribution, i.e., $P^{\gD_l}_Y  = \E_{\Phi_{1:m}}[P^{\gD_l}_{Y|\Phi_{1:m}}]\approx \frac {1}{n}\sum_{i=1}^{n} \sum_{j=1}^m \vw_j P^{s_j}_{Y|\Phi_j}(\cdot|\Phi_j=\bm{\phi}^{s_j}_i)$ should be diverse and close to a uniform distribution. Notice that minimizing the KL divergence between the label distribution and a uniform distribution is equivalent to its entropy maximization. 

Based on these intuitions, we propose to learn the weights $\vw$ by optimizing the information maximization loss~\citep{gomes2010discriminative, hu2017learning},
\begin{align} \label{eq:domain-weights}
\vw^* &= \argmin_{\vw}  \frac {1}{n}\sum_{i=1}^{n} \mathrm{H}\Big(\sum_{j=1}^m \vw_j P^{s_j}_{Y|\Phi_j}(\cdot|\Phi_j=\bm{\phi}^{s_j}_i)\Big) \nonumber \\
&-  \mathrm{H}\Big(\frac {1}{n}\sum_{i=1}^{n}\sum_{j=1}^m \vw_j P^{s_j}_{Y|\Phi_j}(\cdot|\Phi_j=\bm{\phi}^{s_j}_i)\Big),
\end{align}
where $\mathrm{H}$ denotes the entropy function.

\subsection{Surrogate Selection Rule} \label{sec:surrogate}

The key to improving pseudo-labeling is to balance the trade-off between the second bias term and the variance term by only assigning pseudo-labels to a subset of the data $\gD_{l}\triangleq\{(\vx_i^t,\tilde{y}_i^t )\}_{i=1}^{n_l}$ using the selective oracle. However, the oracle, which ensures that the labeled subset $\gD_l$ contains only correctly pseudo-labeled data, is not available in practice. 
To overcome this, we propose a simple yet effective selection rule as the surrogate. Specifically, we adopt a pseudo-label denoising trick to improve the quality of labeling distribution and a confidence query strategy to select new data for pseudo-labeling.

\paragraph{Pseudo-Label Denoising}
Pseudo-labeling techniques are widely used in semi-supervised learning~\citep{lee2013pseudo, shi2018transductive, rizve2021defense}. However, the predictions made by the pretrained source models on target data can be very noisy, so it is crucial to combine them with other labeling criteria to improve the quality of the pseudo-labels. Inspired by \citep{zhang2021prototypical}, we propose a prototype-based pseudo-label denoising method.

For each target sample $\vx^t_i\in \gD_t$, its pseudo-label $\tilde{y}_i^t$ is generated based on two different  criteria: (1) the mixture label distribution directly obtained from source models, i.e., $\vh\left(\{\vf^{s_j}(\vx) \}_{j=1}^m  \right) = \sum_{j=1}^m \vw^*_j P^{s_j}_{Y|\Phi_j}(\cdot|\Phi_j=\bm{\phi}_j)$, and (2) the mixture label distribution obtained using the clustering structure in the representation space, i.e., $\sum_{j=1}^m \vw_j q^{s_j}(\vx_i^t)$, where the label distribution $[q^{s_j}_1,\cdots,q^{s_j}_K]^\top$ is obtained using the distance to prototypes in feature space by ignoring the classifier $\vg^{s_j}$, i.e.,
\begin{align}\label{eq: proto-q}
q^{s_j}_k (\vx_i^t) = \frac{\exp \left(-\left\|\bm{\phi}_i^{s_j}-\bm{\eta}^{s_j}_k\right\| / \tau\right)}{\sum_{k^{\prime}=1}^{K} \exp \left(-\left\|\bm{\phi}_i^{s_j}-\bm{\eta}^{s_j}_{k^\prime}\right\| / \tau\right)}.
\end{align}
Where $\tau$ is the softmax temperature, and $\bm{\eta}^{s_j}_k$ is the prototype of $k$-th class computed using samples from $\gD_l$, i.e.,
\begin{align}\label{eq: proto-eta}
\bm{\eta}^{s_j}_k=\frac{\sum_{\vx_i^t \in \gD_l}\bm{\phi}_i^{s_j} \cdot \mathds{1}\left\{\tilde{y}_i^t=k\right\}}{\sum_{\vx_i^t \in \gD_l} \mathds{1}\left\{\tilde{y}_i^t=k\right\}}.
\end{align}
Assume the two labeling criterion are independent with each other, the probability that they agree on the same pseudo-label $k$ is the product of these two mixture distributions, i.e., $p_k(\vx_i^t) = \big(\sum_{j=1}^m \vw^*_j h^{s_j}_k(\vx_i^t) \big)\cdot 
 \big(\sum_{j=1}^m \vw^*_j q^{s_j}_k(\vx_i^t)\big)$, and we can generate the pseudo-label accordingly, i.e.,
\begin{align} \label{eq:pseudo-label}
\tilde{y}_i^t =  \argmax_{k} p_k(\vx_i^t).
\end{align}
\paragraph{Confidence Query Strategy}
The quantity $ p_k(\vx_i^t)$ can be interpreted as the confidence score of class $k$. The high confidence score implies the generated pseudo-label is more likely to be correct. Thus, we initialize the labeled subset $\gD_l$ with $N$ data with the highest confidence score and denote the remaining unlabeled subset as $\gD_u$. Formally, we partition $\gD^t$ into $\gD_l$ and $\gD_u$ based on the following selection rule,
\begin{equation} \label{eq:split}
\begin{cases}
\left(\vx_i^t,\tilde{y}_i^t \right)  \in \gD_l,   &\text{if } \max_k p_k(\vx_i^t)\geq \alpha_N,\\
 \vx_i^t\in\gD_u,  &\text{Otherwise.}
\end{cases}
\end{equation}
Here, $\alpha_N$ denotes the $N$-th largest confidence score. As shown in Algorithm~\ref{alg:overall}, we query more confident data into $\gD_l$ by increasing $N$ at each iteration of training until $N=n$. More details about the confidence query strategy can be found in Appendix~\ref{sec: app-lambda_alpha}.

\subsection{Self-Training with Data Augmentation}
The labeled subset $\gD_l$ enables us to train the target model in a self-training manner. Notice that the variance term in Theorem~\ref{thm:gen_bound} depends on the number of data in $\gD_l$, i.e., $n_l = |\gX_{\gD_l}|$. Besides querying more pseudo-labeled data as mentioned above, we can further reduce the variance by leveraging data augmentation technique mixup~\citep{zhang2017mixup} to enlarge the subset $\gD_l$, i.e., we construct an augmented dataset $\tilde{\gD_l}$,
\begin{align} \label{eq:mixup}
\tilde{\gD_l} =& \gD_l \cup \{\left(\vx_{\text{aug}}^t,\tilde{y}^t_{\text{aug}}\right)| \vx_{\text{aug}}^t = \lambda \vx_i^t + (1-\lambda)\vx_j^t,  \\
&\tilde{y}^t_{\text{aug}} = \lambda\tilde{y}_i^t+(1-\lambda)\tilde{y}_j^t\, \ ; \left(\vx_i^t, \tilde{y}_i^t\right),\left(\vx_j^t, \tilde{y}_j^t\right) \in \gD_l\},\nonumber
\end{align}
where $\lambda\in [0,1]$ is the mixup ratio.
Since we cannot access the source training data, we fix the source model classifier $\vg^{s_j}$ and update each feature extractor $\vf^{s_j}$ to adapt the target domain, which implicitly incorporates the information from the source domain. 
Therefore, the cross-entropy loss for each feature extractor $\vf^{s_j}$ is given by
\begin{equation*}
\gL_{\text{ce}}\Big(\vf^{s_j}, \tilde{\gD_l} 
\Big)
=-\frac{1}{n_l}\sum_{\vx_i^t,\tilde{y}_i^t\in \tilde{\gD_l}} \sum_{k=1}^K \mathds{1}\left\{\tilde{y}_i^t=k\right\} \log h_k^{s_j}\left(\vx_i^t\right).
\end{equation*}
Moreover, motivated by the same intuition of learning domain aggregation weights in \ref{sec:aggregation}, we also utilize the information maximization loss to encourage the source models to make individual certain but globally diverse predictions, i.e.,

\begin{align}
\gL_{\text{IM}}\Big(\vf^{s_j}, \tilde{\gD_l}
\Big) &= \frac{1}{n_l}\sum_{\vx_i^t\in \tilde{\gD_l}}  \mathrm{H}\left( \vh^{s_j}(\vx_i^t) \right) 
- \mathrm{H}\left( \overline{\vp}^{s_j} \right),
\end{align}
where $\overline{p}_k^{s_j} \triangleq \frac{1}{n_l}\sum_{\vx_i^t\in \tilde{\gD_l}}  h_k^{s_j}\left(\vx_i^t\right)$. 
Different from previous works~\citep{liang2020we, ahmed2021unsupervised}, we only apply information maximization loss on $\tilde{\gD_l}$ to avoid over-confident but wrong pseudo labels in $\gD_u$. Thus, the self-training objectives on $\tilde{\gD_l}$ is given as $\gL_{\tilde{\gD_l}} = \gL_{\text{ce}} + \gL_{\text{IM}}$.

\begin{algorithm}[!t]
\SetAlgoLined
    \textbf{input}{ pretrained source models $\{\vh^{s_j}=\vg^{s_j} \circ \vf^{s_j}\}_{j=1}^{m}$, target domain unlabeled data $\gD^{t}=\{\vx^t_i\}_{i=1}^{n}$, and maximum iterations $T$}

Initialize the domain aggregation weights $\vw^0_j$ by optimizing \eqref{eq:domain-weights} using pretrained source models $\{\vh^{s_j}\}_{j=1}^{m}$.

Initialize the pseudo-label $\tilde{y}_i^t$ for each target data $\vx_i^t$ by~\eqref{eq:pseudo-label} using pretrained models $\{\vh^{s_j}\}_{j=1}^{m}$ and initial domain aggregation weights $\vw^0_j$.

\For{$\tau = 1,2, \ldots, T$}
{   
    Set  $N=N_0+ \frac{(n-N_0)\tau}{T}$.
    
    Update $\gD_l$ and $\gD_u$ by \eqref{eq:split}, and construct the augmented set $\tilde{\gD_l}$ by \eqref{eq:mixup}.

    Update joint discriminators ${\vd^{{(\tau)}}}$ with~\eqref{eq:adv}. 
    
    Update feature extractor $\{\vf^{s_j^{(\tau)}}\}_{j=1}^{m}$ with~\eqref{eq:loss}.
    
    Update the domain aggregation weights $\vw^{\tau}_j$ by optimizing \eqref{eq:domain-weights} using updated models $\{{\vh^{s_j^{(\tau)}}}\}_{j=1}^{m}$.
    
    Update the pseudo-label $\tilde{y}_i^t$ using domain aggregation weights $\vw^{\tau}_j$ and updated models $\{{\vh^{s_j^{(\tau)}}}\}_{j=1}^{m}$ by~\eqref{eq:pseudo-label}.
}

    \textbf{output}{ The final mixture models $\sum_{j=1}^m \vw^{T}_j\vh^{s_j^{(T)}}(\cdot)$.}

 \caption{Selective Self-training for MSFDA}\label{alg:overall}
\end{algorithm}

\subsection{Joint Adversarial Feature Alignment} \label{sec:alignment}
As discussed in~\ref{sec:thm-alignment}, the other essential component of the proposed method is the joint feature alignment. Denote the subset of data not selected for pseudo-labeling as $\gD_u = \gD_t \setminus \gD_l$, then we can align the distribution $P^{\gD_l}_{\Phi_{1:m}}$ and $P^t_{\Phi_{1:m}}$ by enforcing feature alignment between ${\gD_l}$ and $\gD_u$. Intuitively, similar to traditional UDA where the source and target domain have different distributions, we can treat $\gD_l$ as the labeled ``source'' data and $\gD_u$ as the unlabeled ``target'' data, and enforce the joint feature alignment between $\gD_l$ and $\gD_u$.

Specifically, we adopt the adversarial training strategy proposed in~\citep{ganin2015unsupervised, xu2018deep} to perform feature alignment. First, we combine the features extracted from each source model to construct a joint feature representation, i.e., $\{\vf^{s_j}(\vx_i^t)\}_{j=1}^m$. Then, we train a separate neural network $\vd: \mathcal{F}_{1:m} \rightarrow [0,1]$ as the joint discriminator to distinguish the joint features of $\tilde{\gD_l}$ and $\gD_u$, and update multiple feature extractor $\vf^{s_j}$ together to fool the single discriminator. Formally, the joint feature alignment loss is given by the following adversarial loss,
\begin{align}\label{eq:adv}
\gL_{\text{adv}}&\Big(\{\vf^{s_j}\}_{j=1}^m, \vd, \tilde{\gD_l}, \gD_u
\Big)\\\nonumber&
= \frac{1}{n_l}\sum_{\vx_i^t \in \tilde{\gD_l}} \ln \vd(\{\vf^{s_j}(\vx_i^t)\}_{j=1}^m) \\
&+ \frac{1}{n-n_l}\sum_{{\vx_i^t} \in \gD_{u}} \ln \big(1 - \vd(\{\vf^{s_j}(\vx_i^t)\}_{j=1}^m)\big).  \nonumber
\end{align}
In summary, we fix the classifier $\vg^{s_j}$ for each source model, and alternating training the joint discriminator $\vd$ to maximize $\gL_{\text{adv}}$ and training each feature extractor $\vf^{s_j}$ to minimize the combined loss, i.e.,
\begin{equation}
\begin{aligned}
\label{eq:loss}
&\max_{\vd} \gL_{\text{adv}}\Big(\{\vf^{s_j}\}_{j=1}^m, \vd, \tilde{\gD_l}, \gD_u
\Big)\\
&\min_{\{\vf^{s_j}\}_{j=1}^{m}}\sum_{j=1}^{m} \lambda_{\text{ce}}\gL_{\text{ce}}\Big(\vf^{s_j}, \tilde{\gD_l} 
\Big) + \lambda_{\text{IM}}\gL_{\text{IM}}\Big(\vf^{s_j}, \tilde{\gD_l} 
\Big) \\
&+\lambda_{\text{adv}}\gL_{\text{adv}}\Big(\{\vf^{s_j}\}_{j=1}^m, \vd, \tilde{\gD_l}, \gD_u,
\Big),
\end{aligned}
\end{equation}
where $\lambda_{\text{ce}}$, $\lambda_{\text{IM}}$ and $\lambda_{\text{adv}}$ are hyper-parameters that balance different regularization terms.

\subsection{Algorithm}
The overall algorithm is shown in Algorithm \ref{alg:overall}. All the models are retrained using the updated pseudo-labels at each iteration. As the performance of each model increases, we expect to see more correctly labeled samples in $\tilde{\gD_l}$. When the algorithm converges, the final prediction of the target data $\vx^t_i$ is obtained by the mixture models.

\section{Experiment Results}    \label{sec:exp}
\begin{table*}[!t]
  \begin{center}
  \small
  \caption{\textbf{Results on Digit-Five (5 domains):} MN, SV, US, MM, and SY stand for MNIST, SVHN, USPS, MNIST-M, and Synthetic Digits datasets, respectively. Source-ens denotes the ensemble prediction of multiple pretrained source models.}
  \begin{tabular}{llcccccc}
    \\
    \toprule
    \textbf{Setting} & \textbf{Method} & \textbf{MN}& \textbf{SV}& \textbf{US} &\textbf{MM} &\textbf{SY} &\textbf{Avg}\\
    \hline
    Single-source & BAIT~\citep{yang2020unsupervised} & 96.2 & 60.6 & 96.7 & 87.6 & 90.5 & 86.3 \\
              & SFDA~\citep{kim2020domain} & 95.4 & 57.4 & 95.8 & 86.2 & 84.8 & 83.9  \\
              & SHOT~\citep{liang2020we} & 98.9 & 58.3 & 97.7 & 90.4 & 83.9 & 85.8  \\
              & MA~\citep{li2020model} & 98.4 & 59.1 & 98.0 & 90.8 & 84.5 & 86.2\\
    \hline
    Multi-source & Source-ens & 96.7 & 76.8 & 93.8 & 66.7 & 77.6 & 82.3 \\
              & DECISION~\citep{ahmed2021unsupervised} & 99.2 & 82.6 & 97.8 & 93.0 & 97.5 & 94.0  \\
              & CAiDA~\citep{dong2021confident} & 99.1 & 83.3 & \textcolor{violet}{\textbf{98.6}} & 93.7 & 98.1 & 94.6  \\
              & \textbf{Ours} (Surrogate)  & \textcolor{violet}{\textbf{99.2}} & \textcolor{violet}{\textbf{90.7}} & 98.4 &  \textcolor{violet}{\textbf{97.4}} & \textcolor{violet}{\textbf{98.4}} & \textcolor{violet}{\textbf{96.8}} \\
              &  \textbf{Ours} (Selective Oracle)  &\textcolor{red}{99.4} & \textcolor{red}{93.8} & \textcolor{red}{99.0} & \textcolor{red}{98.4} & \textcolor{red}{99.5} & \textcolor{red}{98.0}\\
    \toprule
  \end{tabular}
  \label{table: Digit}
  \end{center}
  \vspace{-1em}
\end{table*}

\begin{table*}[!t]
  \small
  \begin{center}
  \caption{\textbf{Results on Office-31 (3 domains):} A, D, and W stand for Amazon, DSLR, and Webcam datasets, respectively.}
  \begin{tabular}{llcccc}
    \\
    \toprule
    \textbf{Setting} & \textbf{Method} &\textbf{D, W}$\rightarrow$ \textbf{A} & \textbf{A, D}$\rightarrow$ \textbf{W} & \textbf{A, W}$\rightarrow$ \textbf{D} &\textbf{Avg} \\ 
    \hline
    Single-source & BAIT~\citep{yang2020unsupervised} &71.1 &98.5 &98.8 &89.5 \\
              & SFDA~\citep{kim2020domain} &73.2 &93.8 &96.7  &87.9    \\
              & SHOT~\citep{liang2020we} &75.0 &94.9 &97.8  &89.3    \\
              & MA~\citep{li2020model} &75.2 &96.1 &97.3  &89.5 \\
    \hline
    Multi-source  & Source-ens  &62.5 & 95.8 & 98.7 & 86.0  \\
              & DECISION~\citep{ahmed2021unsupervised} & 75.4 & 98.4 & 99.6 & 91.1 \\
              
              & CAiDA~\citep{dong2021confident} &75.8& \textcolor{violet}{\textbf{98.9}} & 99.8 & 91.6  \\
              & \textbf{Ours} (Surrogate) &\textcolor{violet}{\textbf{77.6}} &  98.7 & \textcolor{violet}{\textbf{99.8}} &  \textcolor{violet}{\textbf{92.0}}  \\
              & \textbf{Ours} (Selective Oracle)  &\textcolor{red}{85.5} & \textcolor{red}{99.2} & \textcolor{red}{100.0} & \textcolor{red}{94.9}  \\
    \toprule
  \end{tabular}
  \label{table: Office-31}
  \end{center}
  \vspace{-1em}
\end{table*}

\begin{table*}[!t]
  \small
  \begin{center}
  \caption{\textbf{Results on Office-Home (4 domains):} A, C, R, and P stand for Art, Clipart, Real-world, and Product datasets, respectively.}
  \begin{tabular}{llccccc}
    \\
    \toprule
    \textbf{Setting} & \textbf{Method} & \makecell{\textbf{C, R, P}$\rightarrow$ \textbf{A}} & \makecell{\textbf{A, R, P}$\rightarrow$ \textbf{C}} & \makecell{\textbf{A, C, P}$\rightarrow$ \textbf{R}} &\makecell{\textbf{A, C, R}$\rightarrow$ \textbf{P}}&\textbf{Avg} \\ 
    \hline
    Single-source & BAIT~\citep{yang2020unsupervised}  & 71.1 &59.6 &77.2 &79.4 &71.8 \\
              & SFDA~\citep{kim2020domain}  & 69.3 &57.5 &76.8 &79.1  &70.7  \\
              & SHOT~\citep{liang2020we}  &72.2  &59.3 &82.9 &82.8 &74.3   \\
              & MA~\citep{li2020model} & 72.5 &57.4 &81.7 &82.3  & 73.5\\
    \hline
    Multi-source  & Source-ens  &67.0 &52.1 & 78.6 & 74.8 & 68.1 \\
              & DECISION~\citep{ahmed2021unsupervised} & 74.5 & 59.4 & 83.6 & 84.4 & 75.5\\
              
              & CAiDA~\citep{dong2021confident} & 75.2 & 60.5 & 84.2 & 84.7 & 76.2\\
              & NRC~\citep{yang2021exploiting} &70.6 &60.0 &84.6 &83.5 &74.7\\
              & \textbf{Ours} (Surrogate) &  \textcolor{violet}{\textbf{75.6}} &  \textcolor{violet}{\textbf{62.8}} & \textcolor{violet}{\textbf{84.8}}& \textcolor{violet}{\textbf{85.3}} & \textcolor{violet}{\textbf{77.1}}\\
              & \textbf{Ours} (Selective Oracle)  & \textcolor{red}{86.9} & \textcolor{red}{85.1} & \textcolor{red}{92.2} & \textcolor{red}{95.7} & \textcolor{red}{90.0} \\
    \toprule
  \end{tabular}
  \label{table: Office-Caltech}
  \end{center}
  \vspace{-1em}
\end{table*}

In this section, we describe the experiment settings in~\ref{sec:exp-setting}, present the performance with selective oracle in~\ref{sec:exp-results-oracle} and the results with surrogate selection rule in~\ref{sec:exp-results}, and discuss our takeaways in~\ref{sec: exp-discuss}. Additional experiment results and implementation details can be found in Appendix~\ref{app: exp}.
\subsection{Settings} \label{sec:exp-setting}
\textbf{Datasets}
We conduct extensive evaluations of our methods using the following four benchmark datasets: \textbf{Digits-Five}~\citep{peng2019moment} contains five different domains, including MNIST (MN), SVHN (SV), USPS (US), MNIST-M (MM), and Synthetic Digits (SY). 
\textbf{Office-31}~\citep{saenko2010adapting} contains 31 categories collected from three different office environments, including Amazon(A), Webcam (W), and DSLR (D). \textbf{Office-Home}~\citep{venkateswara2017deep} is a more challenging dataset with 65 categories collected from four different office environments, including Art (A), Clipart (C), Real-world (R), and Product (P). \textbf{DomainNet}~\citep{peng2019moment} is so far the largest and most challenging domain adaptation benchmark, which contains about 0.6 million images with 345 categories collected from six different domains, including Clipart (C), Infograph (I), Painting (P), Quickdraw (Q), Real (R), and Sketch (S).

\textbf{Baselines}
To demonstrate the solid empirical performance of our method, we mainly compare it with the recently proposed SOTA multi-source-free domain adaptation method DECISION~\citep{ahmed2021unsupervised}, CAiDA~\citep{dong2021confident} and NRC~\citep{yang2021exploiting}. We provide another baseline by evaluating ensemble prediction accuracy on the target domain using the pretrained source models, and denote it as Source-ens. In addition, we also include several SOTA single-source-free domain adaptation methods, including BAIT~\citep{yang2020unsupervised}, SFDA~\citep{kim2020domain}, SHOT~\citep{liang2020we}, and MA~\citep{li2020model}. 
These single-source-free methods also do not require access to the source data, and we compare their multi-source ensemble results by taking the average of predictions from the multiple retrained source models after adaptation.

\subsection{Results with Selective Oracle} \label{sec:exp-results-oracle}
In this experiment, we have access to the selective oracle in Definition~\ref{def:oracle} for $\gD_l$ and unlabeled set $\gD_u$ selection. The purpose here is to connect our proposed theorem to empirical performance by demonstrating the level of performance improvement that can be achieved for the MSFDA problem with a perfect subset selection. The results on Digits-Five, Office-31, Office-Home, and DomainNet are shown in the corresponding Tables, denoted as ``Ours (Selective Oracle)". Across all the datasets, it can be observed that our method with selective oracle achieves a considerable performance gain, including more than $10\%$ gain on challenging datasets Office-Home and DomainNet. Although these empirical results can only be achieved with the selective oracle, it validates our theoretical analysis and the significance of balancing the bias-variance trade-off in the MSFDA problem.

\subsection{Results with Surrogate Selection Rule} \label{sec:exp-results}
In practice, as selective oracle is not available, we use the proposed surrogate selection rule discussed in~\ref{sec:surrogate}. The results are denoted as ``Ours (Surrogate)". On the most challenging datasets, Office-Home and DomainNet, our proposed method can still outperform all baseline methods in terms of average accuracy. The SOTA method~\cite{dong2021confident} shows strong performance over other baselines, but our proposed method can still outperform it with significant improvements over several domain adaptation tasks.

\textbf{Digits-Five}
Our method achieves a performance gain of $14.5\%$ over the ensemble of pretrained source models on this dataset. In addition, for some challenging tasks where the source model ensemble performs poorly on the target domain, e.g., M-MNIST, our proposed method outperforms the SOTA method by a large margin of $3.7\%$.

\begin{table*}[!t]
  \begin{center}
  \small
  \caption{\textbf{Results on DomainNet (6 domains):} C, I, P, Q, R, and S stand for Clipart,  Infograph, Painting, Quickdraw, Real, and Sketch, respectively.}
  \begin{tabular}{llccccccc}
    \\
    \toprule
    \textbf{Setting} & \textbf{Method} & \textbf{C}& \textbf{I}& \textbf{P} &\textbf{Q} &\textbf{R} &\textbf{S} &\textbf{Avg}\\
    \hline
    Single-source & BAIT~\citep{yang2020unsupervised} & 57.5 &22.8 &54.1 &14.7 &64.6 &49.2 &43.8 \\
              & SFDA~\citep{kim2020domain} & 57.2 &23.6 &55.1 &16.4 &65.5 &47.3 &44.2  \\
              & SHOT~\citep{liang2020we} & 58.6 &25.2 &55.3 &15.3 &70.5 &52.4 &46.2  \\
              & MA~\citep{li2020model} & 56.8 &24.3 &53.5 &15.7 &66.3 &48.1 &44.1\\
    \hline
    Multi-source & Source-ens & 49.4 & 20.8 & 48.3 & 10.6 & 63.8 & 46.4 & 39.9 \\
              & DECISION~\citep{ahmed2021unsupervised} & 61.5 & 21.6 & 54.6 & 18.9 & 67.5 & 51.0 & 45.9  \\
              & CAiDA~\citep{dong2021confident} & 63.6& 20.7 &54.3& 19.3& \textcolor{violet}{\textbf{71.2}}& 51.6 &46.8\\
              & NRC~\citep{yang2021exploiting} &65.8 &\textcolor{violet}{\textbf{24.1}} &56.0 &16.0 &69.2 &53.4 &47.4\\
              & \textbf{Ours} (Surrogate) & \textcolor{violet}{\textbf{66.5}} & 21.6 &  \textcolor{violet}{\textbf{56.7}} & \textcolor{violet}{\textbf{20.4}} & 70.5 & \textcolor{violet}{\textbf{54.4}} &\textcolor{violet}{\textbf{48.4}}\\
              & \textbf{Ours} (Selective Oracle)   &\textcolor{red}{76.5} & \textcolor{red}{32.8} & \textcolor{red}{64.7} & \textcolor{red}{34.6} & \textcolor{red}{77.8} & \textcolor{red}{62.4} & \textcolor{red}{58.1}  \\
    \toprule
  \end{tabular}
  \label{table: DomainNet}
  \end{center}
  \vspace{-1em}
\end{table*}

\begin{table*}[!htb]
  \begin{center}
  \small
  \caption{\textbf{Ablation Study Results on Digit-Five} }
  \begin{tabular}{lcccccc}
    \\
    \toprule
    \textbf{Method} & \textbf{MN}& \textbf{SV}& \textbf{US} &\textbf{MM} &\textbf{SY} &\textbf{Avg}\\
    \hline
    w/o denoise &99.1 &90.2 &98.0 & 97.4 & 98.2& 96.6\\
    w/o Domain Wights &98.9 &88.7 &98.0 &97.1 & 95.4 &95.6\\
    IM All-Data &98.3 &90.4 &97.5 &96.9 &96.0 &95.8\\
    w/o alignment &99.1 &90.4 & 98.4 &97.3 &98.4&96.7\\
    w/o mixup &99.1 &89.9 & 98.3 &96.9 &97.4 &96.3\\
    \textbf{Ours} (Surrogate)  & \textcolor{violet}{\textbf{99.2}} & \textcolor{violet}{\textbf{90.7}} & \textcolor{violet}{\textbf{98.4}} &  \textcolor{violet}{\textbf{97.4}} & \textcolor{violet}{\textbf{98.4}} & \textcolor{violet}{\textbf{96.8}} \\
                 
    \toprule
  \end{tabular}
  \label{table: Ablation-digit}
  \end{center}
  \vspace{-1em}
\end{table*}

\begin{table*}[!t]
  \small
  \begin{center}
  \caption{\textbf{Ablation Study Results on on Office-31}}
  \vspace{-1em}
  \begin{tabular}{lccccc}
    \\
    \toprule
    \textbf{Method} &\makecell{\textbf{D, W}$\rightarrow$ \textbf{A}} & \makecell{\textbf{A, D}$\rightarrow$ \textbf{W}} & \makecell{\textbf{A, W}$\rightarrow$ \textbf{D}} &\textbf{Avg} \\ 
    \hline
     w/o denoise &76.1 &98.5&99.8&91.5\\
     w/o Domain Wights &76.4 &97.9&99.8&91.4\\
     IM All-Data &74.4 &98.0&99.4&90.6\\
     w/o alignment &77.2 &98.4&99.8&91.8\\
     w/o mixup &76.6 &\textcolor{violet}{\textbf{98.9}}&99.8&91.8\\
     \textbf{Ours} (Surrogate) &\textcolor{violet}{\textbf{77.6}} &  98.7 & \textcolor{violet}{\textbf{99.8}} &  \textcolor{violet}{\textbf{92.0}}  \\
    \toprule
  \end{tabular}
  \label{table: Ablation-office}
  \end{center}
  \vspace{-1em}
\end{table*}

\textbf{Office-31}
It can be seen from Table~\ref{table: Office-31} that most baseline methods perform very well on the tasks A, W $\rightarrow$ D and A, D $\rightarrow$ W, which implies that Domain D and W are similar. Moreover, our proposed method can still exhibit a further improvement on the D, W $\rightarrow$ A task.

\textbf{Office-Home}
This large dataset is more challenging than the Office-31 dataset. For the most difficult task of this dataset: A, R, P $\rightarrow$ C, our approach significantly outperforms the SOTA method~\citep{dong2021confident} by $2.3\%$.

\textbf{DomainNet}
This is the most challenging domain adaptation benchmark so far. Our method still shows superior performance over all baseline methods. 

\subsection{Ablation Study} \label{sec: exp-discuss}
In order to quantify the contributions of each technique discussed in Section~\ref{sec:method}: the learned domain weights, the pseudo-label denoising method,  IM loss evaluated only on $\gD_l$, the joint feature alignment loss $\gL_{\text{adv}}$, and the mixup data augmentation, we take the Digit-Five and Office-31 datasets as examples to perform the ablation study. Specifically, we evaluate the performance of the following variants of our proposed method: (1) w/o Domain Weights, using uniform weights instead of our proposed domain weights. (2) w/o denoise, removing the pseudo-label denoising method.  (3) IM All-Data, apply IM loss on all target domain data instead of only applying on $\gD_l$. (4) w/o alignment, removing the joint feature alignment loss $\gL_{\text{adv}}$. (5) w/o mixup, removing the data augmentation technique mixup, which will induce a larger variance. It can be observed from Table~\ref{table: Ablation-digit} and Table~\ref{table: Ablation-office} that without each of these elements, the performance will degrade. These ablation study results further validate the effectiveness of each of these techniques to balance the bias and variance trade-off discovered by our theorem.

\section{Concluding Remarks}    \label{sec: conclusion}
Our study reveals the significance  of balancing the bias-variance trade-off for the MSFDA problem through information-theoretic analysis, which also provides insights into the algorithm design. The empirical results obtained using a selective oracle assumption demonstrate the performance limit of the MSFDA problem, and it is noteworthy that there is still room for improvement in this source-free setting. As such, identifying a more effective surrogate selection rule remains a promising avenue for future research.

\section*{Acknowledgement}
This work was supported, in part, by the MIT-IBM Watson AI Lab under Agreement No.~W1771646, NSF under Grant No,~CCF-1816209, and ARL and the USAF AI Accelerator under Cooperative Agreement No.~FA8750-19-2-1000. 

\newpage
\bibliography{refs}
\bibliographystyle{icml2023}

\newpage
\appendix
\onecolumn

\section{Derivations \& Proofs}

\subsection{Proof of Theorem~\ref{thm:gen_bound}} \label{appendix: proof}
The gap between the empirical risk $\gL_E(\vh,\gD_l)$ over $\gD_l$ and $\gL_P(\vh,P^t_{XY})$ can be written as
\begin{equation}\label{equ:decomposition}
\begin{aligned}
    &\gL_P(\vh,P^t_{XY}) - \gL_E(\vh,\gD_l)  \\
 &= \gL_P(\vh,P^t_{XY}) -  \gL_P(\vh,P^{\gD_l}_{XY})+ \gL_P(\vh,P^{\gD_l}_{XY})  - \gL_E(\vh,\gD_l) \\
 & = \gL_P(\vh,P^t_{\Phi_{1:m},Y}) -  \gL_P(\vh,P^{\gD_l}_{\Phi_{1:m},Y})+ \gL_P(\vh,P^{\gD_l}_{\Phi_{1:m},Y})  - \gL_E(\vh,\gD_l)\\
 & \le |\gL_P(\vh,P^{\gD_l}_{\Phi_{1:m},Y})  - \gL_E(\vh,\gD_l)|  + |\gL_P(\vh,P^t_{\Phi_{1:m},Y}) -  \gL_P(\vh,P^{\gD_l}_{\Phi_{1:m},Y})|,
\end{aligned}
\end{equation}
where $\gL_P(\vh,P^t_{\Phi_{1:m},Y})\triangleq\E_{P^t_{\Phi_{1:m},Y}} [\ell\left(\vh\left(\Phi_{1:m}  \right), Y\right)]$ explicitly represents $\gL_P(\vh,P^{t}_{XY})$ using $\Phi_{1:m}$. We note that the first term is simply the generalization error of supervised learning  using $n_l$ i.i.d. samples generated from the distribution $P^{\gD_l}_{\Phi_{1:m},Y}$. Since $\vh \in \gH $ has finite Natarajan dimension, by Natarajan dimension theory (see \citep{daniely2011multiclass} equation (6)), the following upper bound holds for some constant $C>0$ with probability at least $1-\delta$,
\begin{equation}\label{equ: VC}
     |\gL_P(\vh,P^{\gD_l}_{\Phi_{1:m},Y})  - \gL_E(\vh,\gD_l)| \le C\sqrt{\frac{d_N(\gH)\log K + \log\frac{1}{\delta} }{n_l}},
\end{equation}
where $K$ is the number of different classes for the label. 

As for the second term in \eqref{equ:decomposition}, it can be upper bounded via the Donsker-Varadhan variational representation of the relative entropy between two probability measures $P$ and $Q$ defined on $\mathcal{X}$:
\begin{equation}
  D(P\|Q) = \sup_{g \in \mathcal G} \Big\{ \mathbb{E}_P[g(X)]  -\log \mathbb{E}_Q[e^{g(X)}] \Big\},
\end{equation}
where the supremum is over all measurable functions $\mathcal G =\{g : \mathcal{X} \to \mathbb{R},\ \mathrm{s.t.}\ \mathbb{E}_Q[e^{g(X)}]< \infty\}$. It then follows that for any $\lambda \in \sR$,
\begin{equation}  D(P^{\gD_l}_{\Phi_{1:m},Y}\|P^t_{\Phi_{1:m},Y}) \ge \mathbb{E}_{P^{\gD_l}_{\Phi_{1:m},Y}}[\lambda\ell(\vh(\Phi_{1:m}),Y)] - \log  \mathbb{E}_{P^t_{\Phi_{1:m},Y}}[e^{\lambda\ell(\vh(\Phi_{1:m}),Y)}].
\end{equation}
Since the loss function $\ell$ is bounded between $[0,M]$,  we can show that $\ell(\vh(\Phi_{1:m}),Y)$ is $\frac{M}{2}$-sub-Gaussian, i.e., 
\begin{equation}
    \log \mathbb{E}_{P^t_{\Phi_{1:m},Y}}\left[e^{\lambda(\ell(\vh(\Phi_{1:m}),Y) - \mathbb{E}_{P^t_{\Phi_{1:m},Y}}[\ell(\vh(\Phi_{1:m}),Y)] )}\right] \le \frac{M^2\lambda^2}{8}.
\end{equation}
Thus, the following inequality holds for all $\lambda \in \sR$,
\begin{equation}
\begin{aligned}
    &D(P^{\gD_l}_{\Phi_{1:m},Y}\|P^t_{\Phi_{1:m},Y}) \\
    &\ge \mathbb{E}_{P^{\gD_l}_{\Phi_{1:m},Y}}[\lambda\ell(\vh(\Phi_{1:m}),Y)] - \log  \mathbb{E}_{P^t_{\Phi_{1:m},Y}}[e^{\lambda\ell(\vh(\Phi_{1:m}),Y)}] \\
    &\ge \lambda \left( \mathbb{E}_{P^{\gD_l}_{\Phi_{1:m},Y}}[\ell(\vh(\Phi_{1:m}),Y)] -  \mathbb{E}_{P^t_{\Phi_{1:m},Y}}[\ell(\vh(\Phi_{1:m}),Y)] \right) - \frac{M^2\lambda^2}{8} \\
    & = \lambda \left( \gL_P(\vh,P^{\gD_l}_{\Phi_{1:m},Y})-\gL_P(\vh,P^t_{\Phi_{1:m},Y})    \right) - \frac{M^2\lambda^2}{8},
    \end{aligned}
\end{equation}
which gives a non-negative parabola in $\lambda$, whose discriminant must be non-positive, which implies
\begin{align}\label{equ: bias_KL}
     \big|\gL_P(\vh,P^t_{\Phi_{1:m},Y}) -  \gL_P(\vh,P^{\gD_l}_{\Phi_{1:m},Y})\big| &\le \sqrt{\frac{M^2}{2} D(P^{\gD_l}_{\Phi_{1:m},Y}\|P^t_{\Phi_{1:m},Y})} \nn \\
     &= \frac{M}{\sqrt{2}}\sqrt{ D(P^{\gD_l}_{\Phi_{1:m}}\|P^t_{\Phi_{1:m}}) +D(P^{\gD_l}_{Y|\Phi_{1:m}}\|P^t_{Y|\Phi_{1:m}}| P^{\gD_l}_{\Phi_{1:m}})}.
\end{align}

Combining \eqref{equ: bias_KL} with \eqref{equ: VC} completes the proof. 

\subsection{Proof of Proposition~\ref{proposition:mixture}} \label{appendix:proposition}
Let $\gP_{w} = \{q~|~ q=\sum_{j=1}^m \vw_j P^{s_j}_{Y|\Phi_j}, \ \sum_{j=1}^m \vw_j=1\}$ denotes the set of mixture distribution. It is easy to show set $\gP_{w}$ is convex and closed. There are two cases, i.e.,  whether the target distribution lies within this convex set or not,
\begin{itemize}
     \item If $P^{t}_{Y|\Phi_{j=1:m}} \in \gP_{w}$, then there exists a $\vw^*$ such that 
    $D\left( \sum_{j=1}^m \vw^*_j P^{s_j}_{Y|\Phi_j} \| P^{t}_{Y|\Phi_{j=1:m}}\right)=0$, and this proposition is trivilly true.
    \item If $P^{t}_{Y|\Phi_{j=1:m}} \notin \gP_{w}$, let $\sum_{j=1}^m \vw^*_j P^{s_j}_{Y|\Phi_j}$ to be the I-projection of $P^{t}_{Y|\Phi_{j=1:m}}$ onto $\gP_{w}$, then by the Pythagoras’ theorem of information geometry (see~\citep{cover1999elements} Theorem 12.6.1), we have
    \begin{equation}
    D\Big(P^{s_j}_{Y|\Phi_j} \| P^{t}_{Y|\Phi_{j=1:m}}\Big) \geq
    D\Big( P^{s_j}_{Y|\Phi_j} \| \sum_{j=1}^m \vw^*_j P^{s_j}_{Y|\Phi_j}\Big)
    +D\Big( \sum_{j=1}^m \vw^*_j P^{s_j}_{Y|\Phi_j} \| P^{t}_{Y|\Phi_{j=1:m}}\Big),
    \end{equation}
    as $P^{s_j}_{Y|\Phi_j} \in \gP_{w}$. Thus, it follows that
    \begin{equation}
    D\Big( \sum_{j=1}^m \vw^*_j P^{s_j}_{Y|\Phi_j} \| P^{t}_{Y|\Phi_{j=1:m}}\Big) \leq D\left(P^{s_j}_{Y|\Phi_j}  \| P^{t}_{Y|\Phi_{j=1:m}}\right).
    \end{equation}
\end{itemize}

\subsection{Proof of Remark~\ref{remark:joint-align}} \label{appendix:remark}
Following the chain rule of KL divergence, we have $D(P^{\gD_l}_{\Phi_{1:m}}\|P^t_{\Phi_{1:m}}) = D(P^{\gD_l}_{\Phi_{j}}\|P^t_{\Phi_{j}}) + D(P^{\gD_l}_{\Phi_{\{1:m\}\setminus j} |\Phi_{j}}\|P^t_{\Phi_{\{1:m\}\setminus j}|\Phi_{j}})$ holds for any $j=1:m$, which implies $D(P^{\gD_l}_{\Phi_{1:m}}\|P^t_{\Phi_{1:m}}) \geq  D(P^{\gD_l}_{\Phi_{j}}\|P^t_{\Phi_{j}})$ holds for any $j$, following by the non-negativity of KL divergence. Therefore, we can conclude that $D(P^{\gD_l}_{\Phi_{1:m}}\|P^t_{\Phi_{1:m}}) \geq \frac 1 m \sum_{j=1}^{m} D(P^{\gD_l}_{\Phi_j}\|P^t_{\Phi_{j}})$.

\section{Additional Experiment Results}\label{app: exp}
\subsection{Visualization}
\begin{figure}[ht]
  \centering
  \captionsetup{font=small}
  \includegraphics[width=0.84\linewidth]{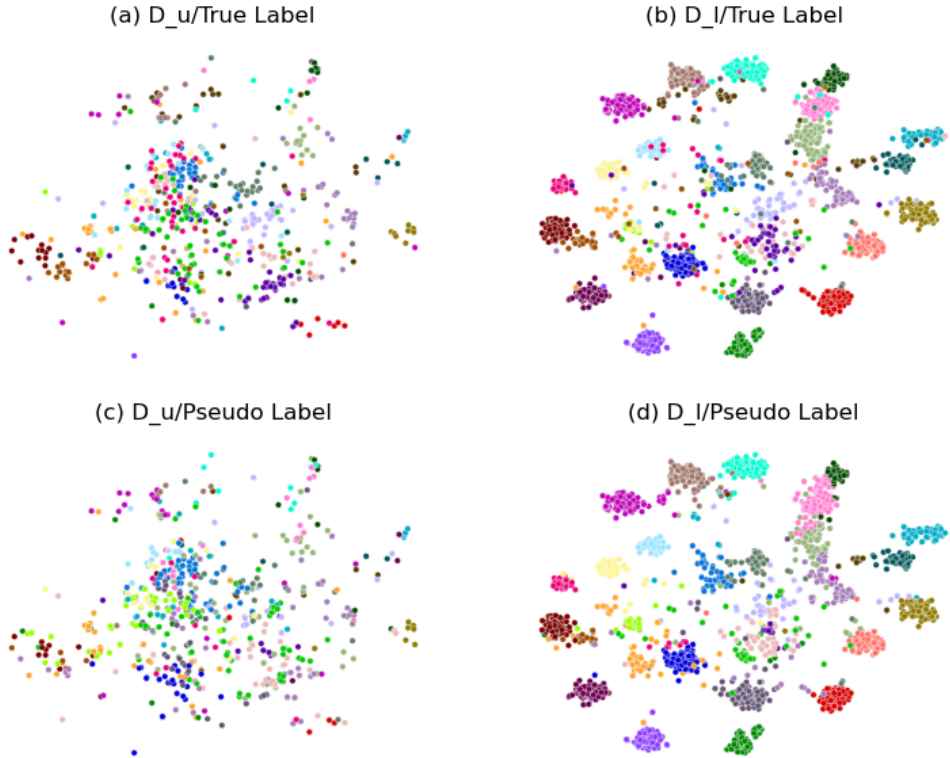}
  \vspace{-1em}
  \caption{t-SNE visualization on D, W $\rightarrow$ A domain adaptation task of Office-31 dataset: (a) plot of $\gD_u$ (705 data) labeled with ground-truth label. (b) plot of $\gD_l$ (2112 data) labeled with the ground-truth label. (c) plot of $\gD_u$ (705 data) labeled with pseudo-label. (d) plot of $\gD_l$ (2112 data) labeled with pseudo-label.}
  \label{fig: t-SNE2}
  \vspace{-1em}
\end{figure}

To better understand why it is crucial to use the selective pseudo labeling technique and joint feature alignment to reduce the two bias terms, we provide t-SNE plots in feature space to visualize the difference between $\gD_l$ and $\gD_u$, and each of them is labeled in two ways: using ground-truth labels and using pseudo labels directly generated from source models. We take task D, W $\rightarrow$ A of the Office-31 dataset as our example and extract the target domain features from pretrained source models. The results are presented in Figure~\ref{fig: t-SNE2}. From the figure, we can find that: (1) There are mismatches between pseudo labels of $\gD_u$ and ground truth labels, i.e., the pseudo labels of $\gD_u$ are much noisier than $\gD_l$. Thus, we should only use data in $D_l$ for training to reduce the first bias term. (2) The feature of samples in $\gD_l$ are more separately clustered compared to $\gD_u$, which implies the existence of a distribution shift between $\gD_l$ and $\gD_u$. To mitigate this discrepancy, we align the feature distributions for both $\gD_l$ and $\gD_u$ by minimizing the joint adversarial feature alignment loss $\gL_{\text{adv}}$ in the proposed method.

\subsection{Implementation Details} \label{sec: app-hyperparameter}
For a fair comparison, we follow the experiment settings in previous works~\citep{ahmed2021unsupervised, liang2020we}. For the Digits-Five benchmark, we use a variant of LeNet~\citep{lecun1998gradient} as our pretrained model. We resize the image samples from different digit domains to the same size ($32\times 32$) and convert the gray-scale images to RGB. For all office benchmarks and DomainNet, we use the pretrained ResNet-50~\citep{he2016deep} as the backbone of the feature extractor, followed by a bottleneck layer with batch normalization and a classifier layer with weight normalization. We train the models on different source domain datasets and then retrain the pretrained models on the remaining single target domain dataset. The weights of the classifier are frozen during model training. The maximum number of training iterations is set to 20. All experiments are implemented in PyTorch using Tesla V100 GPUs with 32 GB memory.

\subsection{Details about Query Strategy}  \label{sec: app-lambda_alpha}
The query strategy in~\eqref{eq:split} relies on a confidence threshold $\alpha_N$ to select data in $\gD_l$ and $\gD_u$. $\alpha_N$ is defined as $N$-th largest confidence score among all the data's confidence score, where $N$ is the number of data to be labeled. The initial number for labeled data $N_0$ is set to be the number of data whose confidence score is larger than an initial threshold, i.e., $\lambda_{\alpha}\cdot \overline{p}$, where $\overline{p}$ denotes the mean confidence score across all the data, and $\lambda_{\alpha}$ is a hyper-parameter. We conduct an ablation study to investigate the impact of $\lambda_{\alpha}$ using Office-31 and Digit-Five dataset, the results are showing in Table~\ref{table: lambda_alpha-office} and Table~\ref{table: lambda_alpha-Digit}. Empirically, we do not observe significant variance of performance with different choices of $\lambda_{\alpha}$. Intuitively, larger $\lambda_{\alpha}$ selects a smaller subset (variance dominates), and smaller $\lambda_{\alpha}$ selects more samples (bias dominates). A balance of bias and variance trade-off is crucial in achieving good performance.

\begin{table}[!t]
  \small
  \captionsetup{font=small}
  \begin{center}
  \caption{\textbf{The impact of $\lambda_{\alpha}$ on Office-31 (3 domains):} A, D, and W stand for Amazon, DSLR, and Webcam datasets, respectively.}
  \begin{tabular}{lcccc}
    \\
    \toprule
    \textbf{Method} &\makecell{\textbf{D,W}$\rightarrow$ \textbf{A}} & \makecell{\textbf{A,D}$\rightarrow$ \textbf{W}} & \makecell{\textbf{A,W}$\rightarrow$ \textbf{D}} &\textbf{Avg} \\ 
    \hline
    $\lambda_{\alpha}=0.5$ &77.8 & 98.2 & 99.8 & 91.9 \\
    $\lambda_{\alpha}=0.6$ &77.6 & 98.7 & 99.8 & 92.0 \\
    $\lambda_{\alpha}=0.8$ & 76.4 & 98.4 &100.0 & 91.6\\
    $\lambda_{\alpha}=1.0$ & 75.9 & 97.7 &100.0 & 91.2\\
    \toprule
  \end{tabular}
  \label{table: lambda_alpha-office}
  \end{center}
\end{table}

\begin{table*}[!t]
  \begin{center}
  \small
  \captionsetup{font=small}
  \caption{\textbf{The impact of $\lambda_{\alpha}$ on Digit-Five (5 domains):} MN, SV, US, MM, and SY stand for MNIST, SVHN, USPS, MNIST-M, and Synthetic Digits datasets, respectively. }
  \begin{tabular}{lcccccc}
    \\
    \toprule
   \textbf{Method} & \textbf{MN}& \textbf{SV}& \textbf{US} &\textbf{MM} &\textbf{SY} &\textbf{Avg}\\
    \hline
    $\lambda_{\alpha}=0.5$  & 99.2 & 90.7 & 98.4 &  97.4 & 98.4 & 96.8 \\
    $\lambda_{\alpha}=0.7$  & 99.1 & 90.4 & 98.6 &  97.3 & 98.4 & 96.8 \\
    $\lambda_{\alpha}=1.0$  & 98.8 & 90.6 & 98.7 &  96.9 & 97.1 & 96.4 \\
    \toprule
  \end{tabular}
  \label{table: lambda_alpha-Digit}
  \end{center}
\end{table*}

\subsection{Hyper-parameter} For model optimization, we use SGD optimizer with momentum value $0.9$ and weight decay $10^{-3}$, the learning rate for backbone, bottleneck layer, and classifier layer is set to $10^{-2}$, $10^{-2}$ and $10^{-3}$, respectively. For domain aggregation weights optimization, we also use SGD optimizer with learning rate $10^{-1}$ without weight decay. The other hyper-parameters are summarized in Table~\ref{table: hyper-parameter}, where $bs$ denotes the batch size, and $itr$ denotes the training iteration.
\begin{table}[h]
  \caption{Hyper-parameter}
  \centering
  \begin{tabular}{llllllllll}
    \toprule
    Task  & $bs$ & $itr$ & $\lambda_{\text{ce}}$ & $\lambda_{\text{IM}}$ & $\lambda_{\text{adv}}$ & $\lambda_{\alpha}$\\
    \midrule
    Office-31 (all domains)  & 32 & 20& 0.1 & 1.0 & 1.0 & 0.6 \\
    \midrule
    Digit-Five (all domains)  & 64 & 20& 0.2 & 1.0 & 1.0 & 0.5 \\
    \midrule
    Office-Home (domain A \& C)  & 32 & 10& 0.2 & 1.0 & 1.0 &0.7 \\
    \midrule
    Office-Home (domain R)  & 32 & 10& 0.2 & 1.0 & 1.0 &0.4 \\
    \midrule
    Office-Home (domain P)  & 32 & 10& 0.2 & 1.0 & 1.0 &0.3 \\
    \midrule
    DomainNet (domain C \& P)  & 32 & 10& 0.2 & 1.0 & 1.0 & 0.5 \\
    \midrule
    DomainNet (domain I, R, S)  & 32 & 10& 0.2 & 1.0 & 1.0 & 0.7 \\
    \midrule
    DomainNet (domain Q)  & 32 & 10& 0.2 & 1.0 & 1.0 & 1.0 \\
    \bottomrule
  \end{tabular}
  \label{table: hyper-parameter}
\end{table}


\end{document}